\documentclass[runningheads]{llncs}
\usepackage{graphicx}

\usepackage{comment}
\usepackage{amsmath,amssymb} 
\usepackage{color}

\usepackage[colorlinks,
            linkcolor=red,       
            citecolor=green,        
            ]{hyperref}
\usepackage[accsupp]{axessibility}  

\usepackage[accsupp]{axessibility}  

\usepackage{multirow}
\usepackage{booktabs}
\usepackage[super]{nth}
\usepackage{makecell}

\begin{document}
\pagestyle{headings}
\mainmatter
\def\ECCVSubNumber{2369}  

\title{Improving RGB-D Point Cloud Registration by Learning Multi-scale Local Linear Transformation} 

\titlerunning{RGB-D Point Cloud Registration by Multi-scale LLT}
%

\author{Ziming Wang\inst{1}$^*$ \and
Xiaoliang Huo\inst{1}$^*$ \and
Zhenghao Chen\inst{2} \and 
Jing Zhang\inst{1} \and
Lu Sheng\inst{1}$^\dagger$ \and
Dong Xu\inst{3}}
\authorrunning{Wang et al.}
%
\institute{School of Software, Beihang University\and
School of Electrical and Information Engineering, The University of Sydney \and
Department of Computer Science, The University of Hong Kong \\ 
\email{\{by1906050,huoxiaoliangchn,lsheng\}@buaa.edu.cn} }

\maketitle
\def\thefootnote{*}\footnotetext{indicates equal contributions}
\def\thefootnote{$\dagger$}\footnotetext{Lu Sheng is the corresponding author, e-mail: \url{lsheng@buaa.edu.cn}}
\setcounter{footnote}{0}
\renewcommand{\thefootnote}{\arabic{footnote}}

\begin{abstract}
Point cloud registration aims at estimating the geometric transformation between two point cloud scans, in which point-wise correspondence estimation is the key to its success.
In addition to previous methods that seek correspondences by hand-crafted or learnt geometric features, recent point cloud registration methods have tried to apply RGB-D data to achieve more accurate correspondence.
However, it is not trivial to effectively fuse the geometric and visual information from these two distinctive modalities, especially for the registration problem.
In this work, we propose a new Geometry-Aware Visual Feature Extractor (GAVE) that employs multi-scale local linear transformation to progressively fuse these two modalities, where the geometric features from the depth data act as the geometry-dependent convolution kernels to transform the visual features from the RGB data.
The resultant visual-geometric features are in canonical feature spaces with alleviated visual dissimilarity caused by geometric changes, by which more reliable correspondence can be achieved.
The proposed GAVE module can be readily plugged into recent RGB-D point cloud registration framework.
Extensive experiments on 3D Match and ScanNet demonstrate that our method outperforms the state-of-the-art point cloud registration methods even without correspondence or pose supervision. The code is available at \href{https://github.com/514DNA/LLT}{https://github.com/514DNA/LLT}.

\keywords{Point cloud registration, geometric-visual feature extractor, local linear transformation}
\end{abstract}

\section{Introduction}
\label{sec:introduction}

Point cloud registration~\cite{el2021unsupervisedr,1992A,choy2020deep,gojcic2020learning,bai2021pointdsc,huang2021predator,ao2021spinnet,yu2021cofinet} is a task to estimate geometric transformation, such as rotation and translation, between two point clouds.
By applying the geometric transformation, we can merge the partial scans from two views of the same 3D scene or object into a complete 3D point cloud, which is a key component of numerous tasks in the community of robotics and AR/VR and also plays an essential role on understanding the whole environment.

The common approach to point cloud registration relies on two processes: (1) correspondence extraction and (2) geometric model fitting, where accurate correspondence is the key for reliable model fitting.
The recent 3D deep learning techniques~\cite{choy20194d,choy2019fully,choy2020deep,el2021unsupervisedr,el2021bootstrap,gojcic2020learning,zeng20173dmatch,2018PPFNet} outperform the traditional methods~\cite{1992A,rister2017volumetric} by finding more accurate correspondence based on learnable geometric features~\cite{choy2019fully,el2021bootstrap}, or further combining the model fitting process into an end-to-end learning framework~\cite{gojcic2020learning,el2021unsupervisedr,el2021bootstrap,choy2020deep}.
However, the geometric features from 3D points are still less discriminative in comparison to visual features from the RGB images.
Thanks to the rapid popularization of RGB-D cameras, it becomes promising to collect the RGB-D data for extracting more reliable correspondence, such that both geometric and visual consistencies can be well examined between two views.
A couple of learning based works~\cite{el2021unsupervisedr,el2021bootstrap} belong to this line of work, which achieve superior registration performance even without ground-truth poses or correspondence as their supervision information.
However, UR\&R~\cite{el2021unsupervisedr} just uses RGB images for correspondence estimation, while BYOC~\cite{el2021bootstrap} relies on pseudo-correspondence from RGB images to train the geometric correspondence. Thus both methods~\cite{el2021unsupervisedr,el2021bootstrap} do not fully leverage the complementary visual and geometric information.
Moreover, according to our experiments (see Section~\ref{sec:experiments}), we can only achieve marginal gains by simply concatenating RGB-D data as the input for correspondence estimation in UR\&R~\cite{el2021unsupervisedr}.
A possible explanation that it is hard to fully exploit the geometry clues by using the CNN networks due to the intrinsic difference between the geometric and visual features.

To this end, we propose a Geometry-Aware Visual Feature Extractor (GAVE) that can generate distinctive but comprehensive geometric-visual features from RGB-D images, which facilitates reliable correspondence estimation for better point cloud registration.
This module can be readily used to replace the feature extractor in UR\&R~\cite{el2021unsupervisedr}, and significantly improve the point cloud registration performance even trained in an unsupervised manner\footnote{As shown in the ablation study, GAVE module can also be applied into the supervised pipelines.}.
To be specific, in the GAVE module, we propose a Local Linear Transformation (LLT) module, where the geometric features (extracted from the geometric feature extractor) act as the guided signal and are converted as point-wise linear coefficients to enhance the visual features (extracted from the visual feature extractor), through point-wise linear transformation.
Moreover, to enhance the content awareness of the transformation with respect to the input depth image, we borrow the idea from the edge-aware image enhancement method~\cite{gharbi2017deep}, which employs the Bilateral Grid and an edge-aware guidance map (both are estimated from the depth image) to generate our content-aware linear coefficients.
Note that this LLT module is applied in the GAVE module in a multi-scale fashion, which thus enriches the scale awareness of the generated visual-geometric features.

More specifically, the proposed LLT module can be viewed as multi-scale dynamic convolutions over visual features that are guided by the geometric clues, which offers more descriptive and complementary combination between visual and geometric features than the common used operations such as concatenation, summation or product.
Since the goemetric feature can represent local geometric structure and indicate local orientations, it is easier for dynamic convolution-based fusion network to learn how to better project visual features into the new feature space where the projected features are robust to geometric changes, which is crucial for registration.
To our best knowledge, it is the first work that applies a dynamic convolution-based fusion strategy in RGB-D point cloud registration, whose design is tailored to the nature of this particular task. 

Our Geometry-Aware Visual Feature Extractor is trained in an end-to-end manner together with the subsequent correspondence estimation and differentiable geometric model fitting modules, \emph{e.g.}, those from UR\&R~\cite{el2021unsupervisedr}.
The state-of-the-art results are achieved on the standard point cloud registration benchmark dataset ScanNet~\cite{dai2017scannet} with the models respectively trained based on the ScanNet~\cite{dai2017scannet} and 3D Match~\cite{zeng20173dmatch} datasets, which clearly outperform the existing point cloud-based supervised baselines and RGB-D-based unsupervised methods.

\section{Related Work}
\label{sec:related_work}

\subsection{3D Feature Extractors.}
To extract the useful 3D features for various 3D vision tasks, early methods adopted the hand-crafted statistic-based strategies~\cite{1992A,2004Distinctive,2006SURF,2011ORB} to discover local 3D geometries.
With the recent success of deep learning techniques, many learning-based 3D feature extraction methods~\cite{2018PPFNet,el2021unsupervisedr,el2021bootstrap,choy20194d,choy2019fully,chen2022exploiting} have been proposed.
While some of them are proposed for extracting the features from point clouds~\cite{choy20194d,choy2019fully,2017PointNet,aoki2019pointnetlk}, our methods are inspired more from those methods that extract the features from RGB-D images/videos~\cite{dai2017scannet,zeng20173dmatch,Silberman:ECCV12,sturm12iros,2015SUN}.
However, most existing geometric-visual feature extractors just simply combine the features respectively from RGB images and depth maps without carefully considering how to exploit their correlation.

\subsection{Bilateral Feature Fusion}

Several methods~\cite{hui2014depth,gharbi2017deep,2017Deep,xia2020joint,xu2021bilateral,chen2022lsvc} have conducted feature fusion in a bilateral manner. Particularly, the works~\cite{gharbi2017deep,xia2020joint,xu2021bilateral} produce the edge-aware affine color transformation by using the Bilateral Grid.
Inspired by those methods, we also develop the content-aware local linear coefficients through the Bilateral Grid, which act as the geometry-guided convolution kernels to transform the visual features.
 
\subsection{3D Point Cloud Registration}

The earlier 3D point cloud registration methods extracted point cloud features and then align them with robust model fitting technologies~\cite{2004PCA,derpanis2004harris,2004Distinctive,2006SURF,2011ORB,detone2018superpoint,choy2019fully}.
Some learning-based methods~\cite{choy2019fully,gojcic2020learning,choy2020deep} leverage the extra ground-truth poses to learn better geometric features from point clouds.
However, it is not trivial to collect such ground-truth annotations.
Recently, the unsupervised learning methods, such as UR\&R~\cite{el2021unsupervisedr} and BYOC~\cite{el2021bootstrap}, enforce cross-view geometric and visual consistency to implicitly supervise the training of registration.
But the features used for correspondence extraction are either directly based on the RGB data~\cite{el2021unsupervisedr} or trained by the pseudo-correspondence labels from the visual correspondences~\cite{el2021unsupervisedr}, where the extraction of visual and geometric clues are usually independent without effectively exploiting their correlation.
Our work is inspired by~\cite{el2021unsupervisedr,el2021bootstrap}, but would like to explore more reliable geometry-aware visual features for more robust registration.
In contrast to those existing methods adopting the fusion operations such as concatenation, summation and attention~\cite{vaswani2017attention}, our newly proposed LLT module explores the correlation between visual and geometric information by using multi-scale dynamic convolutions over visual features, whose kernels are guided by the geometric clues.

\section{Methodology}
\label{sec:methodology}

\begin{figure}[ht!]
   \centering
   \includegraphics[width=\linewidth]{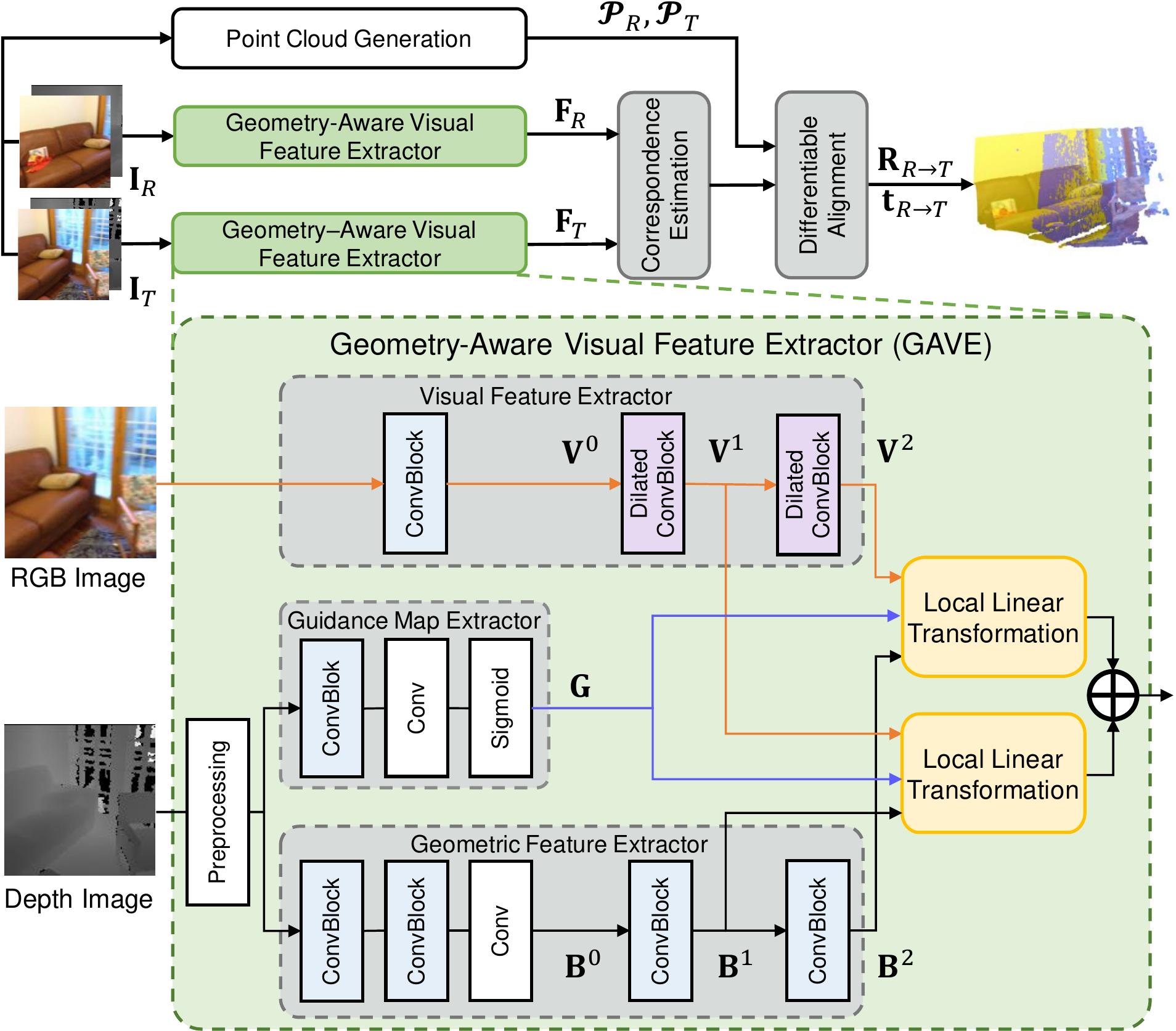}
   \caption{The overview of our Geometry-Aware Visual Feature Extractor (GAVE) based framework. We first generate the multi-scale visual features, multi-scale geometric features and the guidance map, respectively. Then, we fuse the extracted visual and geometric features by using our proposed Local Linear Transformation (LLT) modules with the learned guidance map to produce the intermediate visual-geometric features. The intermediate features from two different scales are then averaged to generate the final visual-geometric features. Once we obtain the pair of the visual-geometric features ($\mathbf{F}_R$, $\mathbf{F}_T$) from the reference RGB-D image and the target RGB-D image, we can then perform the matching and registration operations to produce the rotation matrix and the translation vector by using the correspondence generation and the differentiable alignment module in~\cite{el2021unsupervisedr}. The details of our proposed LLT module and the basic $ConvBlock$ and $Dilated\,ConvBlock$ modules will be illustrated in Figure~\ref{fig:network2}.}
   \label{fig:network}
\end{figure}

In this work, we propose a Geometry-Aware Visual Feature Extractor (GAVE) to learn distinctive and comprehensive geometric-visual features.
Specifically, given each RGB-D image, the GAVE module extracts the visual features and geometric features in a parallel way, and then we densely apply the newly proposed Local Linear Transformation (LLT) module in a multi-scale fashion to progressively fuse the features from these two modalities. 
Therefore, a pair of RGB-D images $\{\mathbf{I}_R, \mathbf{I}_T \}$ ($\mathbf{I}_R$ as the reference RGB-D image, $\mathbf{I}_T$ as the target RGB-D image) can be encoded as a pair of geometric-visual features $\{\mathbf{F}_R, \mathbf{F}_T\}$, which are then inputted into a correspondence estimation module to calculate the correspondence.
The set of captured correspondence is finally used in a geometric model fitting module (\emph{e.g.}, a differentiable alignment module in UR\&R~\cite{el2021unsupervisedr}), together with the point clouds $\mathcal{P}_R$ and $\mathcal{P}_T$ converted from $\mathbf{I}_R$ and $\mathbf{I}_T$, to produce the rotation matrix $\mathbf{R}_{R\rightarrow T} \in \mathbb{R}^{3\times3}$ and the translation vector $\mathbf{t}_{R\rightarrow T} \in \mathbb{R}^{3\times1}$ from the reference RGB-D image to the target RGB-D image.
In our work, we adopt the correspondence estimation and differentiable alignment modules from UR\&R~\cite{el2021unsupervisedr} in addition to our GAVE module, thus the whole registration framework can be trained in an end-to-end unsupervised learning manner.
The overall framework is shown in Figure~\ref{fig:network}.

\subsection{Overview of Our Geometry-Aware Visual Feature Extractor}

In our Geometry-Aware Visual feature Extractor, we have two parallel sub-networks, namely a visual feature extractor and a geometric feature extractor respectively, to extract the visual and geometric features from a RGB-D image.
The visual feature extractor contains $2$ dilated convolution blocks to enlarge the receptive fields that describes the visual contents.
The geometric features after the last $2$ convolution blocks are converted into a set of Bilateral Grids~\cite{gharbi2017deep}, which act as the source of the local linear coefficients for the proposed Local Linear Transformation modules.
In addition, based on depth image, we also produce an edge-aware guidance map that further helps to interpolate geometry-dependent linear coefficients from the predicted Bilateral Grids.
Since then, the LLT module progressively applies the extracted guidance map to slice the set of Bilateral Grids, by which the resultant local linear coefficients are employed to transform the visual features, which are extracted after the last two dilated convolutional blocks in the visual feature extractor.
Since our GAVE module adopts a multi-scale fusion strategy, we can produce the final visual-geometric features by averaging the outputs from both LLT modules.
More details of each component will be respectively introduced in the following sections.

\subsection{Visual Feature Extractor}
\label{Visual Feature}

\begin{figure}[t]
   \includegraphics[width=\linewidth]{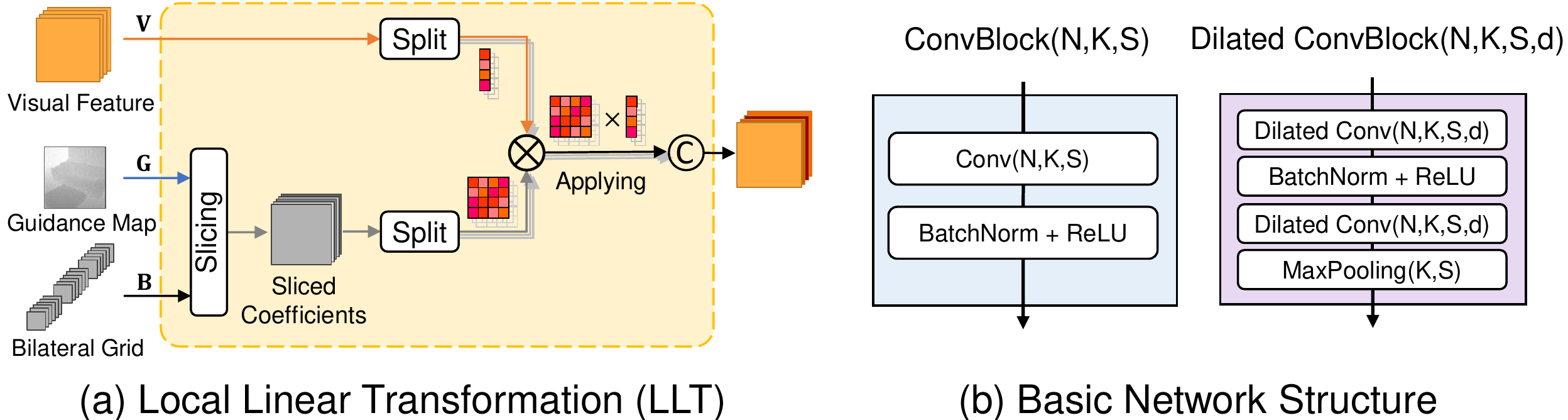}
   \caption{
   For our ``Local Linear Transformation'' in (a), we first produce the sliced coefficients from the guidance map and Bilateral Grid by using the slicing operation in~\cite{gharbi2017deep}.
   Then, at each position, the sliced coefficient matrix will be split in to group-wise coefficient matrix.
   To produce the transformed feature at each group, we then perform the applying operation (\textit{i.e.,} the linear transformation) between each group-wise coefficient and each group-wise visual feature, which is also split from the previously learned visual feature.
   Finally, we generate the final output feature by using the channel-wise concatenation operation on these transformed features from all groups.
   The basic module is shown in (b) ``Basic Network Structure''. ``$Conv(N,K,S)$” represents
    the convolution operation with the output channel, the kernel size and the stride as $N$,
    $K \times K$ and $S$, respectively. ``d'' in ``$Dilated\,Conv(N,K,S,d)$” refers to the dilation parameter of the dilated convolution operation.}
   \label{fig:network2}
\end{figure}

We apply dilated convolutions to enlarge the receptive fields in the visual feature extractor.
Specifically, the visual feature extractor at first extracts an initial visual feature map $\mathbf{V}^0 \in{\mathbb{R}^{H\times W \times D_c}}$ by using a $ConvBlock(64, 3, 1)$ operation (\textit{i.e.,} $D_c=64$). $H$ and $W$ are the height and width of the input RGB image.
Then, two dilated convolution blocks $Dilated\,ConvBlock(64,3,1,2)$ are stacked thereafter, where visual feature maps $\mathbf{V}^1\in{\mathbb{R}^{H\times W \times D_c}}$ and $\mathbf{V}^2\in{\mathbb{R}^{H\times W \times D_c}}$ are generated from each block, as the sources for multi-scale feature fusion.
There are no downsampling operations in this module, thus the output visual feature maps have the same spatial size as the input image.
Note that the definitions of $ConvBlock(N, K, S)$ and $Dilated\,ConvBlock(N, K, S, d)$ are depicted in Figure~\ref{fig:network2}(b), where $N$ is the number of output channel, $K$ is the kernel size, $S$ is the stride and $d$ refers to the dilation factor.

\subsection{Geometric Feature Extractor}
\label{Content-aware}

The input depth image is at first normalized to $[0,1)$ through some linear normalization operations with a \texttt{sigmoid} function.
Since raw depth image may contain holes due to sensor's systematic errors, thus in the pre-processing step, we also apply the Joint Bilateral Filtering (JBF) method~\cite{kopf2007joint} to fill the depth holes with the aid of the corresponding RGB image.

Once we produce the normalized depth map, we first encode it by using a stack of convolution operations (\textit{i.e.,} $ConvBlock(32, 3, 2)$, $ConvBlock(256, 3, 2)$ and $Conv(768, 3, 2)$\footnote{$Conv(N, K, S)$ is a standard convolution operation with the output channel $N$, the kernel size $K$ and the stride size $S$.}) to generate an initial down-scaled geometric feature map $\mathbf{B}^0\in{\mathbb{R}^{(H/8)\times (W/8) \times D_d}}$ (we use $D_d=768$ in this work, so as to match the size of the visual features, which will be explained in Section~\ref{Local Linear}).
Since then, we use two $ConvBlock(768, 3, 1)$ modules to respectively generate two geometric feature maps representing two different scales.
Each Bilateral Grid is reshaped from each geometric feature map as $\mathbf{B}^i \in{\mathbb{R}^{ (H/8)\times (W/8) \times (D_d/n_{grid}) \times n_{grid}}}$, where $i = 1, 2$, and $n_{grid}$ is the depth of the Bilateral Grid (in this work we set $n_{grid}=3$ for balancing the efficiency and effectiveness).
Please refer to~\cite{gharbi2017deep} for more details about Bilateral Grid.

\subsection{Guidance Map Extractor}
\label{Edge-aware}

In order to provide the content-structural information when generating the local linear coefficients, we define a guidance map by using a point-wise nonlinear transformation on the depth map.
Specifically, we input the normalized and hole-filled depth map and then directly employ several convolution operations (\textit{i.e.,} $ConvBlock(3, 3, 1)$ and $Conv(1, 3, 1)$) and a \texttt{sigmoid} activation function to produce a learned guidance map $\mathbf{G} \in {\mathbb{R}^{H\times W \times 1}}$, which preserves the piece-wise smoothness as well as discontinuity presented in the depth image.

\subsection{Multi-scale Local Linear Transformation}
\label{Local Linear}

In order to effectively fuse the visual and geometric clues, we progressively apply the local linear transformation to learn the visual-geometric features in a multi-scale manner.
To be specific, in each LLT module, we would like to \emph{slice} the generated Bilateral Grids to produce the content-aware local linear coefficients, and then \emph{apply} the sliced coefficients to transform the visual features.
For the sake of efficiency, we also split the local linear transformation into several groups evenly along the channel dimension, and then concatenate these group-wise outputs as the final visual-geometric features.

\subsubsection{Slicing}
The slicing operation is performed between a Bilateral Grid $\mathbf{B}$ (\textit{i.e.,} $\mathbf{B}^1$ or $\mathbf{B}^2$) and the guidance map $\mathbf{G}$.
At each spatial location in the guidance map, we use its spatial coordinates and the value at that location to sample nearest points in the Bilateral Grid, and then bilinearly interpolate the sampled coefficients to eventually generate the sliced linear coefficients.
Therefore, the sliced coefficients become a tensor $\tilde{\mathbf{A}} \in \mathbb{R}^{H\times W \times (D_d/n_{grid})}$.
Note that we split the local linear transformation into $n_{group}$ ($n_{group} = 16$ in this work) groups, namely the sliced coefficients $\tilde{\mathbf{A}}$ can be reshaped as $\mathbf{A} \in \mathbb{R}^{H\times W \times D_c \times(D_c/n_{group})}$.
Note that when $D_c = 64$ and $D_d = 768$, $n_{grid} = 3$ and $n_{group} = 16$, $\tilde{\mathbf{A}}$ and ${\mathbf{A}}$ have the same number of elements.
In this way, the context of the sliced coefficient tensor will be conditioned on the content structure from the guidance map, and such slicing operation is more computational-friendly than \texttt{softmax}-based interpolation. 

\subsubsection{Apply}

After evenly splitting the sliced linear coefficients $\mathbf{A}$ to produce $\mathbf{A}^{(g)} \in \mathbb{R}^{H\times W \times (D_c/n_{group}) \times (D_c/n_{group})}, g = 1, \ldots, n_{group}$, the final local linear transformation can be obtained by first using a point-wise transformation and then using a channel-wise concatenation, such as
\begin{equation}
    \mathbf{F} = \overset{n_{group}} {\underset{g=1}{\|}} \mathbf{A}^{(g)} \otimes \mathbf{V}^{(g)}
\end{equation}
where $\overset{n_{group}} {\underset{g=1}{\|}}$ means channel-wise concatenation among $n_{group}$ linearly transformed group-wise features, $\otimes$ indicates the matrix multiplication operation at every spatial position.
$\mathbf{V}^{(g)} \in \mathbb{R}^{H\times W \times (D_c/n_{group})}$ is the $g$-th group of $\mathbf{V}$, which is evenly split from $\mathbf{V}$ along the channel dimension.
Moreover, $\mathbf{F} \in \mathbb{R}^{H\times W \times D_c}$ has the same size as the visual feature $\mathbf{V}$.
Note that we take the LLT module from one scale as an example for better illustration. We omit the superscript $i$ (\textit{i.e.,} the scale index) in $\mathbf{V}$, $\mathbf{V}^{(g)}$ and $\mathbf{F}$ for brevity, as the LLT module from each scale shares the same process. 
Thus, our local linear transformation-based fusion method inherently takes advantage of both modalities and provide a more flexible fusion strategy than simple concatenation or summation operations.

\subsubsection{Multi-scale Fusion}

The fused visual-geometric features $\mathbf{F}^i \in \mathbb{R}^{H\times W \times D_c}, i = 1,2$ in both scales are then averaged at the end of the GAVE module, so as to fulfill the multi-scale awareness of the features, which is essential for correspondence estimation in point cloud registration.

\subsection{Correspondence, Registration and Objective Functions}
\label{Matching}

\subsubsection{Correspondence and Registration}

After feeding two RGB-D images $(\mathbf{I}_R, \mathbf{I}_T)$ to our GAVE extractor to produce the visual-geometric feature pairs $(\mathbf{F}_R$, $\mathbf{F}_T)$, we can then perform the following correspondence estimation and differentiable alignment operations.
Specifically, we first follow the work in~\cite{el2021unsupervisedr} to compute the top-$k$ (we set $k=400$ in this work) correspondence pairs and then employ such correspondence pairs to estimate the rotation matrix $\mathbf{R}_{R\rightarrow T}$ and the translation vector $\mathbf{t}_{R\rightarrow T}$ by using the differentiable alignment module~\cite{el2021unsupervisedr}.

\subsubsection{Objective Function}

As proposed by UR\&R~\cite{el2021unsupervisedr}, we also apply the photometric, depth and correspondence consistencies to train the whole RGB-D point cloud registration framework. The photometric consistency is measured by comparing the target image with the differentiably rendered reference image, according to the estimated rotation and translation parameters. The depth consistency is similar to the photometric consistency, but it compares the depth value instead. Correspondence consistency directly measures the matching errors between the corresponded points. Please refer to~\cite{el2021unsupervisedr} for more details.

\section{Experiments}
\label{sec:experiments}

\subsection{Datasets and Experimental Setup}

\subsubsection{Datasets}

We follow UR\&R~\cite{el2021unsupervisedr} and adopt the large-scale indoor RGB-D dataset ScanNet~\cite{dai2017scannet} for evaluating our proposed GAVE module.
Specifically, there are $1513$ scenes in
the ScanNet dataset~\cite{dai2017scannet} and each scene contains both RGB-D images and their ground-truth camera poses. 
We use its original training/testing split, which respectively contain $1045$ and $312$ scenes.
In addition, as in~\cite{el2021unsupervisedr}, we also provide more evaluation results, in which we train our model based on another smaller point cloud dataset 3D Match~\cite{zeng20173dmatch} with $101$ real-world indoor scenes and then evaluate the learnt model on the ScanNet dataset. Each scene in 3D Match also provides RGB-D images and point clouds data.

\begin{table*}[t]
    \centering
    \caption{Pairwise registration errors on the ScanNet~\cite{dai2017scannet} dataset. We report the mean and median errors in terms of rotation error (°), translation error (mm), and Chamfer distance (cm). Features for correspondence estimation may come from visual or geometric/3D modality. The training set can be 3D Match~\cite{zeng20173dmatch} or ScanNet~\cite{dai2017scannet}. ``Sup'' means training with ground-truth pose supervision.
    }
    \renewcommand{\arraystretch}{0.3}
    \scriptsize 
    \begin{tabular}{c|cccc|cccccccc}
        \toprule
        \multicolumn{1}{c}{\multirow{3}{*}{Methods}} &
        
        \multicolumn{1}{c}{\multirow{3}{*}{\makecell[c]{Train Set}}} &
        \multicolumn{1}{c}{\multirow{3}{*}{\makecell[c]{Sup}}}  &
        \multicolumn{2}{c}{Features} &
        \multicolumn{2}{c}{Rotation} & \multicolumn{2}{c}{Translation} & \multicolumn{2}{c}{Chamfer} &
        \multicolumn{1}{c}{\multirow{3}{*}{\makecell[c]{FMR}}}\\

        \cmidrule(r){4-5} \cmidrule(r){6-7} \cmidrule(lr){8-9} \cmidrule(l){10-11} 
        \multicolumn{2}{c}{} &  & Visual & 3D & Mean & Med. & Mean & Med.& Mean & Med. &\\ \midrule
        SIFT~\cite{2004Distinctive} & N/A & & \checkmark & & 18.6 & 4.3 & 26.5  & 11.2 & 42.6 & 1.7 & - \\ \\
        SuperPoint~\cite{detone2018superpoint} & N/A & & \checkmark & & 8.9 & 3.6 & 16.1 & 9.7 & 19.2 & 1.2 & -\\ \\
        FCGF~\cite{choy2019fully} & N/A & & & \checkmark & 9.5 & 3.3 & 23.6 & 8.3 & 24.4 & 0.9 & -\\ 
        \midrule
        
        BYOC~\cite{el2021bootstrap} & 3D Match & & \checkmark & \checkmark & 7.4 & 3.3 & 16.0 & 8.2 & 9.5 & 0.9 & -\\ \\
        DGR~\cite{choy2020deep} & 3D Match & \checkmark & & \checkmark & 9.4 & 1.8 & 18.4 & 4.5 & 13.7 & 0.4 & -\\ \\
        3D MV Reg~\cite{gojcic2020learning} & 3D Match & \checkmark & & \checkmark & 6.0 & 1.2 & 11.7 & 2.9 & 10.2   & 0.2 & -\\ \\
        
        UR\&R~\cite{el2021unsupervisedr} & 3D Match & & \checkmark & & 4.3 &  1.0 &  9.5 &  2.8 & 7.2 &  0.2  & 0.78\\ \\
        UR\&R (RGB-D) & 3D Match & & \checkmark & \checkmark & 3.8 &  1.1 &  8.5 &  3.0 &  6.5 &  0.2   & 0.78\\ \\
        Ours & 3D Match & & \checkmark & \checkmark & \textbf{3.0} & \textbf{0.9} & \textbf{6.4} & \textbf{2.4} &  \textbf{5.3} & \textbf{0.1} & \textbf{0.87}\\


        \midrule
        BYOC~\cite{el2021bootstrap} & ScanNet & & \checkmark & \checkmark & 3.8 & 1.7 & 8.7 & 4.3 & 5.6 & 0.3 & -\\ \\
        UR\&R~\cite{el2021unsupervisedr} & ScanNet & & \checkmark &  & 3.4 & 0.8 & 7.3 & 2.3 & 5.9 & 0.1 & 0.85 \\ \\
        UR\&R (RGB-D) & ScanNet & & \checkmark & \checkmark & 2.6 & 0.8 & 5.9 & 2.3 & 5.0 &  0.1 & 0.91\\ \\
        
        Ours & ScanNet & & \checkmark & \checkmark & \textbf{2.5} & \textbf{0.8} & \textbf{5.5} & \textbf{2.2} &  \textbf{4.6} & \textbf{0.1} & \textbf{0.94} \\
        
        \bottomrule
    \end{tabular} 
    \label{tab:ScanNetResult}%
\end{table*}

\begin{table*}[t]
    \centering
    \caption{Pairwise registration accuracies on the ScanNet~\cite{dai2017scannet} dataset. We report the rotation accuracy with different angles (\textit{i.e.,} 5°, 10° and 45°), the translation accuracy with different lengths (\textit{i.e.,} 5cm, 10cm and 25cm) and the Chamfer accuracy with different metric distances (\textit{i.e.,} 1mm, 5mm and 10mm).
    %
    }
    \renewcommand{\arraystretch}{0.3}
    \scriptsize 
    \begin{tabular}{c|cccc|cccccccccc}
        \toprule
        \multicolumn{1}{c}{\multirow{3}{*}{Methods}} & \multicolumn{1}{c}{\multirow{3}{*}{\makecell[c]{Train Set}}} & 
        \multicolumn{1}{c}{\multirow{3}{*}{\makecell[c]{Sup}}}  &
        \multicolumn{2}{c}{Features} &
        \multicolumn{3}{c}{Rotation} & \multicolumn{3}{c}{Translation} & \multicolumn{3}{c}{Chamfer} \\
          
        \cmidrule(r){4-5} \cmidrule(lr){6-8} \cmidrule(lr){9-11} \cmidrule(l){12-14} 
        
        \multicolumn{2}{c}{} &  & Visual & 3D & 5 & 10 & 45 & 5 & 10 & 25 & 1 & 5 & 10 \\ \midrule
        SIFT~\cite{2004Distinctive} & N/A & & \checkmark & & 55.2 & 75.7  & 89.2 & 17.7 & 44.5 & 79.8 & 38.1  & 70.6  & 78.3 \\ \\
        SuperPoint~\cite{detone2018superpoint} & N/A & & \checkmark & & 65.5 & 86.9 & 96.6 & 21.2  & 51.7 & 88.0 & 45.7  & 81.1 & 88.2 \\ \\
        FCGF~\cite{choy2019fully} & N/A & & & \checkmark & 70.2 & 87.7  & 96.2  & 27.5  & 58.3  & 82.9 & 52.0  & 78.0  & 83.7\\ 
        \midrule
        
        BYOC~\cite{el2021bootstrap} & 3D Match & & \checkmark & \checkmark & 66.5 & 85.2  & 97.8 & 30.7  & 57.6  & 88.9 & 54.1 & 82.8  & 89.5 \\ \\
        DGR~\cite{choy2020deep} & 3D Match & \checkmark & & \checkmark & 81.1 & 89.3  & 94.8 & 54.5  & 76.2  & 88.7 & 70.5  & 85.5  & 89.0 \\ \\
        3D MV Reg~\cite{gojcic2020learning}  & 3D Match & \checkmark & & \checkmark & 87.7 & 93.2 & 97.0 & 69.0  & 83.1  & 91.8 & 78.9  & 89.2  & 91.8 \\ \\
        
        UR\&R~\cite{el2021unsupervisedr} & 3D Match & &\checkmark & & 87.6 & 93.1 & 98.3 & 69.2 & 84.0  & 93.8 & 79.7 & 91.3  & 94.0 \\ \\
        UR\&R (RGB-D)~\cite{el2021unsupervisedr} & 3D Match & & \checkmark & \checkmark & 87.6 & 93.7 & 98.8 & 67.5 & 83.8 & 94.6 & 78.6 & 91.7 & 94.6     \\ \\
        Ours & 3D Match & & \checkmark & \checkmark & \textbf{93.4} & \textbf{96.5} & \textbf{98.8} &  \textbf{76.9} & \textbf{90.2} & \textbf{96.7} &  \textbf{86.4} & \textbf{95.1} & \textbf{96.8} \\


        \midrule
        BYOC~\cite{el2021bootstrap} & ScanNet & & \checkmark & \checkmark & 86.5 & 95.2  & 99.1 & 56.4  & 80.6  & 96.3 & 78.1 & 93.9  & 86.4\\ \\
        UR\&R~\cite{el2021unsupervisedr} & ScanNet & & \checkmark & & 92.7 & 95.8 & 98.5 & 77.2 & 89.6 & 96.1 & 86.0 & 94.6 & 96.1 \\ \\
        UR\&R (RGB-D)~\cite{el2021unsupervisedr} & ScanNet & & \checkmark & \checkmark & 94.1 & 97.0 & 99.1 & 78.4 & 91.1 & 97.3 & 87.3 & 95.6 & 97.2 \\ \\
        
        Ours & ScanNet & & \checkmark & \checkmark & \textbf{95.5} & \textbf{97.6} & \textbf{99.1} & \textbf{80.4} & \textbf{92.2} & \textbf{97.6} & \textbf{88.9} & \textbf{96.4} & \textbf{97.6} \\
        
        \bottomrule
    \end{tabular} 
    \label{tab:ScanNetResult2}%
\end{table*}

\subsubsection{Evaluation Metrics}

We adopt the evaluation metrics, \textit{i.e.,} the rotation error, the translation error and the Chamfer distance, as used in UR\&R~\cite{el2021unsupervisedr}.
We report both the mean and the median values for these three error metrics in Section~\ref{results} and Section~\ref{ablation}. 
In addition, we report the registration accuracy, \textit{i.e.}, the rotation accuracy within three thresholds of angles, the translation accuracy within three thresholds of lengths and the Chamfer accuracy within three thresholds of metric distances, as introduced in UR\&R~\cite{el2021unsupervisedr}.
We also include FMR~\cite{choy2019fully,2018PPFNet} to directly compare the extracted correspondence with the reference methods, in which we use rigorous thresholds $\tau_{1}=0.05$ and $\tau_{2}=0.5$.

\subsubsection{Baseline Methods}

We compare our work with the conventional registration methods, which extract 3D features by SIFT~\cite{2004Distinctive}, SuperPoint~\cite{detone2018superpoint} and FCGF~\cite{choy2019fully}, and then estimate the geometric transformation via RANSAC.
Moreover, we compare with the learning-based registration approaches, such as DGR~\cite{choy2020deep} and 3D MV Reg~\cite{gojcic2020learning} as the supervised approaches, and UR\&R~\cite{el2021unsupervisedr}, BYOC~\cite{el2021bootstrap} as the unsupervised approaches.
The results of these methods are borrowed from~\cite{el2021unsupervisedr,el2021bootstrap}.
Last, we also use the RGB-D images as the input for correspondence estimation in UR\&R~\cite{el2021unsupervisedr} (\textit{i.e.,}~UR\&R~(RGB-D)), as another important baseline method.

\subsubsection{Training Details}

For fair comparison, we follow the same training scheme as in~\cite{el2021unsupervisedr}.
Specifically, we train our model based on the 3D Match dataset for only $14$ epochs with the learning rate of 1e-4.
We also train our model based on the ScanNet dataset for only $1$ epoch with the learning rate 1e-4.
All models are trained on the machine with one NVIDIA Tesla V100 GPU. The batch size is $8$. We use Adam Optimizer~\cite{2014Adam} with epsilon 1e-4 and momentum $0.9$.

\subsection{Experimental Results}
\label{results}

We provide our experimental results, \textit{i.e.,} registration errors in Table~\ref{tab:ScanNetResult} and registration accuracies in Table~\ref{tab:ScanNetResult2}.
It is observed that our newly proposed method not only outperforms UR\&R (RGB-D), but also achieves significant improvement over the baseline methods~\cite{2004Distinctive,detone2018superpoint,choy2019fully,el2021bootstrap,2020Deep,gojcic2020learning,el2021unsupervisedr}.
Specifically, our method trained on the 3D Match dataset achieves much better results than all other end-to-end optimized methods that are also trained on the 3D Match dataset.
For example, when compared to the most recent unsupervised point cloud registration method UR\&R~\cite{el2021unsupervisedr}, we respectively reduce $21.1\%$ mean rotation error, $24.7\%$ mean translation error, and $18.5\%$ mean Chamfer distance.
We also increase the FMR performance for about $11.54\%$, which directly validates the superior correspondence estimation performance of our method.
With respect to registration accuracies, we also achieve significant gains at the strictest thresholds.
These results demonstrate that our proposed network has universal registration ability, because significant gains can be achieved on the large-scale ScanNet dataset by simply training the network in a smaller 3D Match dataset.

We have similar observations when compared with these methods trained on the ScanNet dataset.
But without any domain gap between training \& testing data, the baseline methods can achieve almost saturated performance (over $90\%$ in terms of most metrics for UR\&R). While it is non-trivial to achieve further gains in this case, our method still reduces up to $23.7\%$/$9.3\%$/$12.6\%$ relative error rate over the baselines in terms of rotation/translation/Chamfer distance.

\begin{table*}[t]
    \centering
    \caption{Comparision between our complete method (\textit{i.e.,} the \nth{4} row) and three alternative methods, which directly adopt the concatenation of RGB images and depth maps for generating the intermediate feature.
    ``MS'' means multi-scale strategy, ``DC'' means dilated convolutions in the visual feature extractor, and ``LLT'' is the local linear transformation module.
    All models are trained based on the 3D Match dataset.}
    \renewcommand{\arraystretch}{0.5}
    \scriptsize
    \begin{tabular}{ccc|ccccccccccccccccc}
        \toprule
        \multicolumn{1}{c}{\multirow{4}{*}{\makecell[c]{MS}}} &
        \multicolumn{1}{c}{\multirow{4}{*}{\makecell[c]{DC}}} &
        \multicolumn{1}{c}{\multirow{4}{*}{\makecell[c]{LLT}}} &
        \multicolumn{5}{c}{Rotation} & \multicolumn{5}{c}{Translation} & \multicolumn{5}{c}{Chamfer} \\
        & & &
        \multicolumn{3}{c}{Accuracy} & 
        \multicolumn{2}{c}{Error} & 
        \multicolumn{3}{c}{Accuracy} & 
        \multicolumn{2}{c}{Error} &
        \multicolumn{3}{c}{Accuracy } & 
        \multicolumn{2}{c}{Error}  \\
        \cmidrule(r){4-6} \cmidrule(lr){7-8} \cmidrule(lr){9-11} \cmidrule(lr){12-13} \cmidrule(lr){14-16} \cmidrule(l){17-18} 
        & &  & 5 & 10 &45 &Mean & Med. & 5 & 10&25 & Mean & Med. & 1 & 5 &10& Mean & Med. \\
                \midrule
        & & & 88.4 & 94.2 & 98.6 &  3.8 &  1.1 & 67.3 & 83.8 & 94.5 &  8.5 &  3.0 & 78.9 & 91.7 & 94.6 &  6.5 &  0.2  \\ \midrule
       \checkmark& & &  88.5 & 94.4 & 98.6 &  3.8 &  1.1 & 68.1 & 84.5 & 94.8 &  8.3 &  3.0 & 79.5 & 92.1 & 94.9 &  6.3 &  0.2 \\ \midrule
        \checkmark& \checkmark&  & 90.4 & 95.0 & 98.6 &  3.6 &  1.0 & 70.8 & 86.5 & 95.3 &  8.1 &  2.8 & 81.8 & 93.1 & 95.4 &  6.2 &  0.2      \\ \midrule
       \checkmark& \checkmark & \checkmark&  \textbf{93.4} & \textbf{96.5} & \textbf{98.8} & \textbf{3.0} &  \textbf{0.9} & \textbf{76.9} & \textbf{90.2} &\textbf{96.7} & \textbf{6.4} &  \textbf{2.4} & \textbf{86.4} & \textbf{95.1} & \textbf{96.8} & \textbf{5.3} &  \textbf{0.1} \\ \bottomrule
        
    \end{tabular}
    \label{tab:Ablation}
\end{table*}

\begin{table*}[t]
    \centering
    \caption{Comparision between our complete method and four variants, which are (1) the method without adopting the fusion mechanism at all (\textit{i.e.,} the \nth{3} row in Table~\ref{tab:Ablation}),
    (2) the method without the guidance map, and
    (3) the method that replaces LLT by the affine transformation.
    (4) the method that replaces LLT by multi-head cross-attention (MHCA).
    All models are trained on the 3D Match dataset.}
    \renewcommand{\arraystretch}{0.3}
    \scriptsize
    \begin{tabular}{c|ccccccccccccccc}
        \toprule
        \multicolumn{1}{c}{\multirow{4}{*}{\makecell[c]{Fusion \\ Strategies}}} &

        \multicolumn{5}{c}{Rotation} & \multicolumn{5}{c}{Translation} & \multicolumn{5}{c}{Chamfer} \\
        &
        \multicolumn{3}{c}{Accuracy} & 
        \multicolumn{2}{c}{Error} & 
        \multicolumn{3}{c}{Accuracy} & 
        \multicolumn{2}{c}{Error} &
        \multicolumn{3}{c}{Accuracy } & 
        \multicolumn{2}{c}{Error}  \\
        \cmidrule(r){2-4} \cmidrule(lr){5-6} \cmidrule(lr){7-9} \cmidrule(lr){10-11} \cmidrule(lr){12-14} \cmidrule(l){15-16} 
        &  5 & 10 &45 & Mean & Med. & 5 & 10 & 25&Mean & Med. & 1 & 5 &10 &Mean & Med. \\
                \midrule
         no fusion & 90.4 & 95.0 & 98.6 &  3.6 &  1.0 & 70.8 & 86.5 & 95.3 &  8.1 &  2.8 & 81.8 & 93.1 & 95.4 &  6.2 &  0.2      \\ \midrule
       \makecell[c]{ LLT w/o \\guidance map} & 92.3 & 95.9 & 98.8 &  3.2 &  0.9 & 74.7 & 88.7 & 96.3 &  7.0 &  2.5 & 84.6 & 94.4 & 96.2 &  5.4 &  0.1  \\ \midrule
        \makecell[c]{ Affine \\transformation} & 92.3 & 96.0 & 98.8 &3.1 &  0.9 & 75.3 & 89.0& 96.3& 6.7 &  2.5 & 85.2 & 94.5 & 96.3& 5.4 &  0.1 \\ \midrule
        {MHCA} & {91.3} & {95.1} & {98.4} & {3.8} & {0.9} & {73.5} & {87.6} & {95.2} &  {8.4} &  {2.6} & {83.4} & {93.3} & {95.4} &  {6.5} &  {0.2} \\ \midrule
        LLT (Ours) & \textbf{93.4} & \textbf{96.5} & \textbf{98.8} & \textbf{3.0} &  \textbf{0.9} & \textbf{76.9} & \textbf{90.2} & \textbf{96.7} & \textbf{6.4} &  \textbf{2.4} & \textbf{86.4} & \textbf{95.1} & \textbf{96.8} & \textbf{5.3} &  \textbf{0.1} \\ \bottomrule
        
    \end{tabular}
    \label{tab:Fusion}
\end{table*}

\begin{table}[t]
    \renewcommand{\arraystretch}{0.5}
    \scriptsize
     \caption{
    Comparision between our method and UR\&R (RGB-D) when trained based on ground truth camera poses.
    All models are trained on the 3D Match dataset.}
    \begin{tabular}{cccccccccccccccc}
        \toprule
        \multicolumn{1}{c}{\multirow{4}{*}{\makecell[c]{Methods}}} &
        \multicolumn{5}{c}{Rotation} & \multicolumn{5}{c}{Translation} & \multicolumn{5}{c}{Chamfer} \\
        &
        \multicolumn{3}{c}{Accuracy} & 
        \multicolumn{2}{c}{Error} & 
        \multicolumn{3}{c}{Accuracy} & 
        \multicolumn{2}{c}{Error} &
        \multicolumn{3}{c}{Accuracy } & 
        \multicolumn{2}{c}{Error}  \\
        \cmidrule(r){2-4} \cmidrule(lr){5-6} \cmidrule(lr){7-9} \cmidrule(lr){10-11} \cmidrule(lr){12-14} \cmidrule(l){15-16} 
        & 5 & 10 &45 & Mean & Med. & 5 & 10 & 25&Mean & Med. & 1 & 5 &10 &Mean & Med. \\
     \midrule
      UR\&R (RGB-D)&  92.3  & 95.3  & 98.2  & 3.8   & 0.8  & 77.6  & 89.4  & 95.5  & 7.8   & 2.3   & 86.1  & 94.0 & 95.6  & 6.7   &  0.1  \\ \midrule
      Ours & \textbf{96.5}  & \textbf{97.8} & \textbf{98.8} & \textbf{2.7}  & \textbf{0.8}   & \textbf{83.8} & \textbf{93.8} & \textbf{97.6} & \textbf{5.8}  & \textbf{2.0}   & \textbf{91.2} & \textbf{96.7} & \textbf{97.6} & \textbf{4.8}  & \textbf{0.1} \\ \midrule
    \end{tabular}
    \label{tab:sup}
\end{table}

\subsection{Ablation Study and Analysis} 
\label{ablation}
\subsubsection{Analysis of Each Component}
In Table~\ref{tab:Ablation}, we analyse the effectiveness of each proposed component, \textit{i.e.,} the local linear transformation (LLT) module, the multi-scale (MS) fusion strategy and dilated convolution (DC), by comparing our complete method to three alternative methods. We train all models based on the 3D Match dataset.
The first variant in the \nth{1} row replaces the dilation convolutions with regular convolutions, and does not adopt either multi-scale fusion strategy or the LLT module.
The second variant in the \nth{2} row introduces the multi-scale fusion strategy upon the first alternative, and the third variant in the \nth{3} row further includes dilated convolution in the visual feature extractor.
The first variant achieves the worst registration performance, while the second one reduces $2.4\%$ mean translation error and $3.1\%$ mean Chamfer distance when compared to the first variant.
The third variant further reduces $7.6\%$ mean rotation error, $2.4\%$ mean translation error and $1.6\%$ mean Chamfer distance when compared to the second alternative method.
Note that our complete method in the \nth{4} row after using the LLT module can bring the most significant gains.

\subsubsection{Different Fusion Strategies}

In Table~\ref{tab:Fusion}, we further compare our LLT based fusion module with the other alternatives. We train these models on the 3D Match dataset. The first variant in the \nth{1} row does not adopt the fusion mechanism at all, which achieves the worst registration performance.
The second one in the \nth{2} row directly uses the Bilateral Grid without using the guidance map, while the third variant in the \nth{3} row replaces the linear transformation by the affine transformation. 
It is observed that both the second and the third variants can intuitively bring some performance improvements.
In contrast, our proposed LLT module in the \nth{5} row can bring the most significant gains.
It is interesting that the variant using affine transformation is worse than that using the linear transformation. A possible explanation is that the bias term in the affine transformation indicates another summation operation between the geometric features and the visual features, which may deteriorate the feature representation if two modalities are quite different.
Last, in the \nth{4} row, we adopt the fast multi-head cross attention (MHCA) mechanism of Linear Transformer~\cite{katharopoulos2020transformers} to replace the LLT module. Here, we do not apply vanilla MHCA~\cite{vaswani2017attention} to avoid huge memory and computational costs.
The results show that LLT is better than this variant in terms of all evaluation metrics.

\begin{figure}[t]
   \centering
  \includegraphics[width=0.98\linewidth]{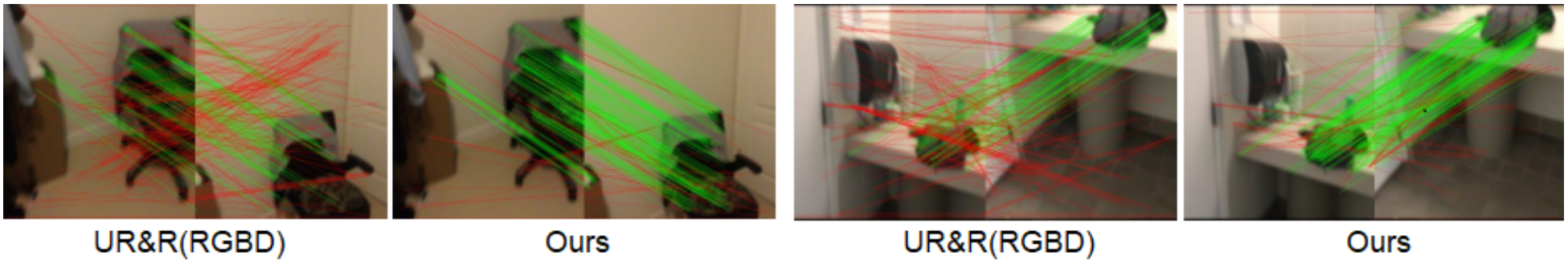}
   \caption{Visualization of pairwise matching results by UR\&R (RGB-D)~\cite{el2021unsupervisedr} and our method (trained on the 3D Match dataset~\cite{zeng20173dmatch}). We show the positive correspondence (\textit{i.e.,} the matching error $<$ 10cm) and the negative correspondence (\textit{i.e.,} the matching error $\geq$ 10cm) as the green lines and the red lines, respectively. Best viewed on screen.} 
   \label{fig:fig4}
\end{figure}

\begin{figure}[t]
   \centering
   \includegraphics[width=0.97\linewidth]{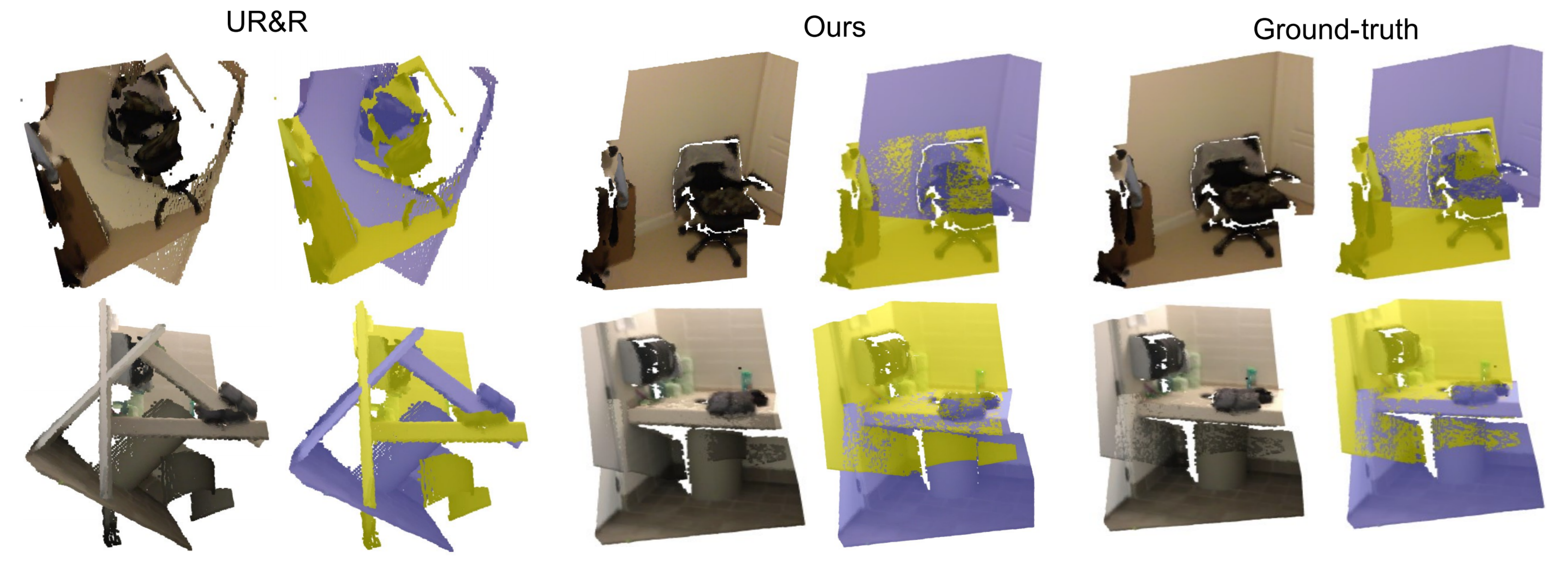}
   \caption{Visualization of point cloud registration results from UR\&R (RGB-D)~\cite{el2021unsupervisedr} and our method (trained on the 3D Match dataset~\cite{zeng20173dmatch}). 
   In the \nth{1}, the \nth{3} and the \nth{5} columns, we show the stitched 3D scenes; while we use the purple and yellow points to represent the point clouds from the target and reference viewpoints (see the \nth{2}, the \nth{4} and the \nth{6} columns). Best viewed on screen.} 
   \label{fig:fig5}
\end{figure}

\subsubsection{Supervised Learning}

In Table~\ref{tab:sup}, our proposed feature extractor can be trained under the supervised learning setting. Specifically, as in~\cite{choy2020deep}, we adopt the camera pose data as the ground-truth labels during the training procedure, for both our method and our baseline UR\&R (RGB-D).
The results show that our method is much better than UR\&R (RGB-D) in terms of all metrics.

 \begin{figure}[t]
   \centering
   \includegraphics[width=1.0\linewidth]{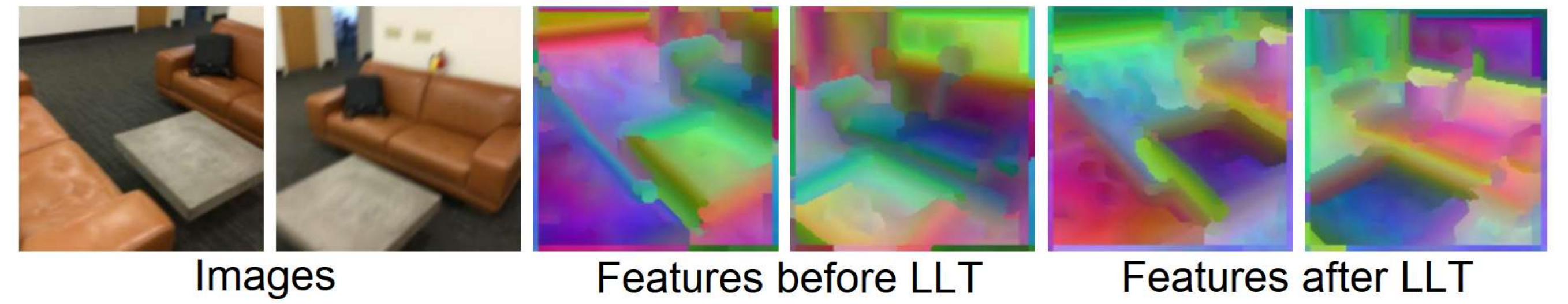}
   \caption{The \nth{3} and the \nth{4} columns represent the projected 3D features, which are the input to our LLT, while the \nth{5} and the \nth{6} columns represent the projected 3D features, which are the output from our LLT. We use t-SNE~\cite{van2008visualizing} for visualization, in which each 3D feature is mapped to the corresponding color.}
   \label{fig:rebuttal_visual}
 \end{figure}
 
\subsection{Qualitative Results}

\subsubsection{Visual Comparison about Correspondence Estimation and Registration}

In Figure~\ref{fig:fig4}, we visualize the matching results for both UR\&R~\cite{el2021unsupervisedr} (RGB-D) and our proposed method, in which we use the models trained on 3D Match.
It is observed that our method provides more accurate matching results across two views.
Taking the results in the left of Figure~\ref{fig:fig4} as an example, UR\&R (RGB-D) finds false correspondence around the plain area within the floor and wall, while our method pays more attention to salient objects, such as the chairs, where the correspondence can be found in a more reliable and repeatable way. 
In Figure~\ref{fig:fig5}, we visualize the registration results for both UR\&R~\cite{el2021unsupervisedr} (RGB-D) and our method.
It is observed that our method achieves better registration results.
For example, in the \nth{1} and \nth{4} rows, our method generates very close stitching results to the ground-truth, while the results from the UR\&R (RGB-D) method are completely failed.
As we can already produce more accurate matching results, it is not surprised that we can achieve better point cloud registration performance than UR\&R (RGB-D). 

\subsubsection{Feature Visualization}
In Figure~\ref{fig:rebuttal_visual}, the \nth{3} and \nth{4} columns and the \nth{5} and \nth{6} columns are the projected 3D features from left and right images by using t-SNE, before and after using the LLT module.
We observe that the learnt features (e.g. within the table area) after using our LLT are more likely to follow the geometric structure, and have become more consistent across two views.

\section{Conclusion}

In this work, we have proposed a new geometric-aware visual feature extractor (GAVE) to effectively learn visual-geometric features, in which we propose multi-scale local linear transformation to progressively fuse the geometric and visual features.
Our proposed GAVE module can be easily plugged into different end-to-end point cloud registration pipelines like~\cite{el2021unsupervisedr} (as already discussed in this work), which significantly enhances the point cloud registration performance. 
Extensive experiments not only show our method outperforms the existing registration methods, but also indicate the effectiveness of our newly proposed LLT module and multi-scale fusion strategy. 
It is possible to further extend and apply our proposed GAVE feature extractor for more RGB-D based 3D computer vision tasks, such as recognition, tracking, reconstruction and \emph{etc.}, which will be studied in our future work.

\subsubsection{Acknowledgement}

This work was partially supported by the National Natural Science
Foundation of China (No. 61906012, No. 62132001, No. 62006012).

\clearpage
%
%
\bibliographystyle{splncs04}
\bibliography{id2369_ref}



\pagestyle{headings}
\def\ECCVSubNumber{2369}  


\clearpage
\title{Improving RGB-D Point Cloud Registration by Learning Multi-scale Local Linear Transformation \\ (Supplementary Material)}


\titlerunning{RGB-D Point Cloud Registration by Multi-scale LLT}
%
\maketitle


\section{Analysis of Hyperparameters}
\renewcommand{\thefigure}{S\arabic{figure}}
\renewcommand{\thetable}{S\arabic{table}}
\begin{table*}[h]
    \centering
    \caption{Results of our method when using different number of grids (\textit{i.e.,} $n_{grid}$). We perform the experiments with the fixed number of groups (\textit{i.e.,} $n_{group}=16$) and various number of grids.
    All models are trained based on the 3D Match dataset and evaluated on the ScanNet dataset.}
    \renewcommand{\arraystretch}{0.5}
    \scriptsize
    \begin{tabular}{c|ccccccccccccccccc}
        \toprule
        \multicolumn{1}{c}{\multirow{4}{*}{\makecell[c]{$n_{grid}$}}} &
        \multicolumn{5}{c}{Rotation} & \multicolumn{5}{c}{Translation} & \multicolumn{5}{c}{Chamfer} \\
        &
        \multicolumn{3}{c}{Accuracy} & 
        \multicolumn{2}{c}{Error} & 
        \multicolumn{3}{c}{Accuracy} & 
        \multicolumn{2}{c}{Error} &
        \multicolumn{3}{c}{Accuracy } & 
        \multicolumn{2}{c}{Error}  \\
        \cmidrule(r){2-4} \cmidrule(lr){5-6} \cmidrule(lr){7-9} \cmidrule(lr){10-11} \cmidrule(lr){12-14} \cmidrule(l){15-16} 
        & 5 & 10 &45 &Mean & Med. & 5 & 10&25 & Mean & Med. & 1 & 5 &10& Mean & Med. \\
                \midrule
        2 & 92.6 & 96.1 & 98.9 &  3.0 &  0.9 & 74.9 & 89.0 & 96.3 &  6.6 &  2.5 & 84.9 & 94.5 & 96.4 &  5.4 &  0.1 \\ \midrule
        3 & 93.4 & 96.5 & 98.8 & 3.0 & 0.9 & 76.9 & 90.2 & 96.7 & 6.4 &  2.4 & 86.4 & 95.1 & 96.8 & 5.3 &  0.1 \\  \midrule
        4 & 93.1 & 96.2 & 98.7 &  3.1 &  0.9 & 75.9 & 89.6 & 96.5 &  6.7 &  2.4 & 85.8 & 94.8 & 96.4 &  5.7 &  0.1    \\\bottomrule
        
    \end{tabular}
    \label{tab:Ablation_supp}
\end{table*}

\begin{table*}[h]
    \centering
    \caption{Results of our method when using different number of groups (\textit{i.e.,} $n_{group}$). We perform the experiments with the fixed number of grids (\textit{i.e.,} $n_{grid}=3$) and various number of groups.
    All models are trained based on the 3D Match dataset and evaluated on the ScanNet dataset.}
    \renewcommand{\arraystretch}{0.5}
    \scriptsize
    \begin{tabular}{c|ccccccccccccccccc}
        \toprule

        \multicolumn{1}{c}{\multirow{4}{*}{\makecell[c]{$n_{group}$}}} &
        \multicolumn{5}{c}{Rotation} & \multicolumn{5}{c}{Translation} & \multicolumn{5}{c}{Chamfer} \\
        &
        \multicolumn{3}{c}{Accuracy} & 
        \multicolumn{2}{c}{Error} & 
        \multicolumn{3}{c}{Accuracy} & 
        \multicolumn{2}{c}{Error} &
        \multicolumn{3}{c}{Accuracy } & 
        \multicolumn{2}{c}{Error}  \\
        \cmidrule(r){2-4} \cmidrule(lr){5-6} \cmidrule(lr){7-9} \cmidrule(lr){10-11} \cmidrule(lr){12-14} \cmidrule(l){15-16} 
        & 5 & 10 &45 &Mean & Med. & 5 & 10&25 & Mean & Med. & 1 & 5 &10& Mean & Med. \\
                \midrule
        32 & 92.5 & 96.1 & 98.8 &  3.1 &  0.9 & 75.4 & 89.0 & 96.3 &  6.7 &  2.5 & 85.1 & 94.6 & 96.4 &  5.4 &  0.1   \\\midrule
        16&   93.4 & 96.5 & 98.8 & 3.0 & 0.9 & 76.9 & 90.2 & 96.7 & 6.4 &  2.4 & 86.4 & 95.1 & 96.8 & 5.3 &  0.1 \\ \midrule
        8 & 92.5 & 96.0 & 98.8 &  3.2 &  0.9 & 75.3 & 89.1 & 96.2 &  6.8 &  2.5 & 85.0 & 94.5 & 96.3 &  5.7 &  0.1     \\
        \bottomrule
        
    \end{tabular}
    \label{tab:Ablation2}
\end{table*}

\begin{table}[h]
    \renewcommand{\arraystretch}{0.5}
    \scriptsize
    \centering
    \caption{\textcolor{black}{Results of our method when using different scales. We compare our complete method (\textit{i.e.,} the \nth{2} row) with two alternative methods, which directly changed the scale to 1 (\textit{i.e.,} the \nth{1} row) and 3 (\textit{i.e.,} the \nth{3} row).
    All models are trained based on the 3D Match dataset and evaluated
on the ScanNet datase.}}
    \begin{tabular}{cccccccccccccccc}
        \toprule
        \multicolumn{1}{c}{\multirow{4}{*}{\makecell[c]{scale}}} &
        \multicolumn{5}{c}{Rotation} & \multicolumn{5}{c}{Translation} & \multicolumn{5}{c}{Chamfer} \\
        &
        \multicolumn{3}{c}{Accuracy} & 
        \multicolumn{2}{c}{Error} & 
        \multicolumn{3}{c}{Accuracy} & 
        \multicolumn{2}{c}{Error} &
        \multicolumn{3}{c}{Accuracy } & 
        \multicolumn{2}{c}{Error}  \\
        \cmidrule(r){2-4} \cmidrule(lr){5-6} \cmidrule(lr){7-9} \cmidrule(lr){10-11} \cmidrule(lr){12-14} \cmidrule(l){15-16} 
        & 5 & 10 &45 & Mean & Med. & 5 & 10 & 25&Mean & Med. & 1 & 5 &10 &Mean & Med. \\
    \midrule

      1 &92.1 & 95.9 & 98.8 &  3.2 &  0.9 & 75.6 & 88.8 & 96.3 &  6.8 &  2.4 & 85.0 & 94.5 & 96.4 &  5.5 &  0.1  \\ \midrule
      2 & 93.4 & 96.5 & 98.8 &  3.0 &  0.9 & 76.9 & 90.2 & 96.7 &  6.4 &  2.4 & 86.4 & 95.1 & 96.8 &  5.3 &  0.1 \\ \midrule
      3&92.0 & 95.9 & 98.9 &  3.1 &  1.0 & 74.1 & 88.4 & 96.2 &  6.8 &  2.6 & 84.0 & 94.3 & 96.3 &  5.4 &  0.1
       \\ \midrule
    \end{tabular}
    \label{tab:scale}
\end{table}

We evaluate the sensitivity of our method with respect to two parameters: the number of grids (\textit{i.e.,} $n_{grid}$) and the number of groups (\textit{i.e.,} $n_{group}$), and the corresponding results are reported in Table~\ref{tab:Ablation_supp} and Table~\ref{tab:Ablation2}.
Specifically, we train our model based on the 3D Match dataset~\cite{zeng20173dmatch} and test our model on the ScanNet dataset~\cite{dai2017scannet}. All the experiments are performed on NVIDIA 2080 Ti GPUs.
Note that we set the number of grids as $3$ and the number of groups as $16$ (\textit{i.e.,} the \nth{2} row in Table~\ref{tab:Ablation_supp} and Table~\ref{tab:Ablation2}) under the default setting in our main paper.

In Table~\ref{tab:Ablation_supp}, it is observed that we can achieve better registration performance when we increase the number of grids from $2$ to $3$. A possible explanation is that our method using more grids allows the guidance map to have more flexible slicing with respect to the grids.
For example, we achieve better rotation accuracy at 5° (\textit{i.e.,} 0.9\% improvement), translation accuracy at 5cm (\textit{i.e.,} 2.7\% improvement) and Chamfer Distance at 1mm (\textit{i.e.,} 2.1\% improvement).
However, we also see the performance of our model using $4$ grids (\textit{i.e.,} $n_{grid}=4$) drops when compared to our model using $3$ grids. Moreover, our model using $4$ grids costs $12.8\%$ extra GPU memory than our model by using $3$ grids.
Hence, we use $3$ grids as our default setting in the main paper.

Similarly, in Table~\ref{tab:Ablation2}, it is observed that we can achieve better registration performance when reducing the number of groups from $32$ to $16$. For example, we can see that  the mean rotation error, the mean translation error and the mean Chamfer Distance from our model when using $32$ groups (\textit{i.e.,} the \nth{1} row) are respectively increased by 3.2\%, 4.5\% and 1.9\% when compared to our model when using $16$ groups (\textit{i.e.,} the \nth{2} row).
While, when we continue reduce the number of groups to $8$ (\textit{i.e.,} the \nth{3} row), we observe the performance drops again when compared to our model using $8$ groups ,namely, the mean rotation error, the mean translation error and the mean Chamfer Distance are respectively increased by 6.7\%, 4.2\% and 7.5\% when compared to our model when using $16$ groups.
It is also worth mentioning that we need to consume $42.5\%$ more peak GPU memory for our model using $8$ groups when compared to our model using $16$ groups.
Hence, we use $16$ groups as our default setting in the main paper.

\textcolor{black}{As for in Table.~\ref{tab:scale}, we analysis our approach’s performance by using different numbers of scales (\textit{i.e.,} 1, 2 and 3). The results indicate that our GAVE with more scales does not necessarily improve the performance yet brings more computational burden and the risk of overfitting. Hence, we choose ``scale=2” as it achieves the best results with an acceptable trade-off.}

The overall results from Table~\ref{tab:Ablation_supp} and Table~\ref{tab:Ablation2} also indicate that the mean rotation, translation and Chamfer errors achieved by our method are relatively stable when setting $n_{grid}$ within the range of [2, 3, 4] and when setting $n_{group}$ within the range of [8, 16, 32].
Hence, we simply choose $n_{grid}=3$ and $n_{group}=16$ based on a relatively better performance-complexity trade-off.

\begin{figure}[t]
   \includegraphics[width=\linewidth]{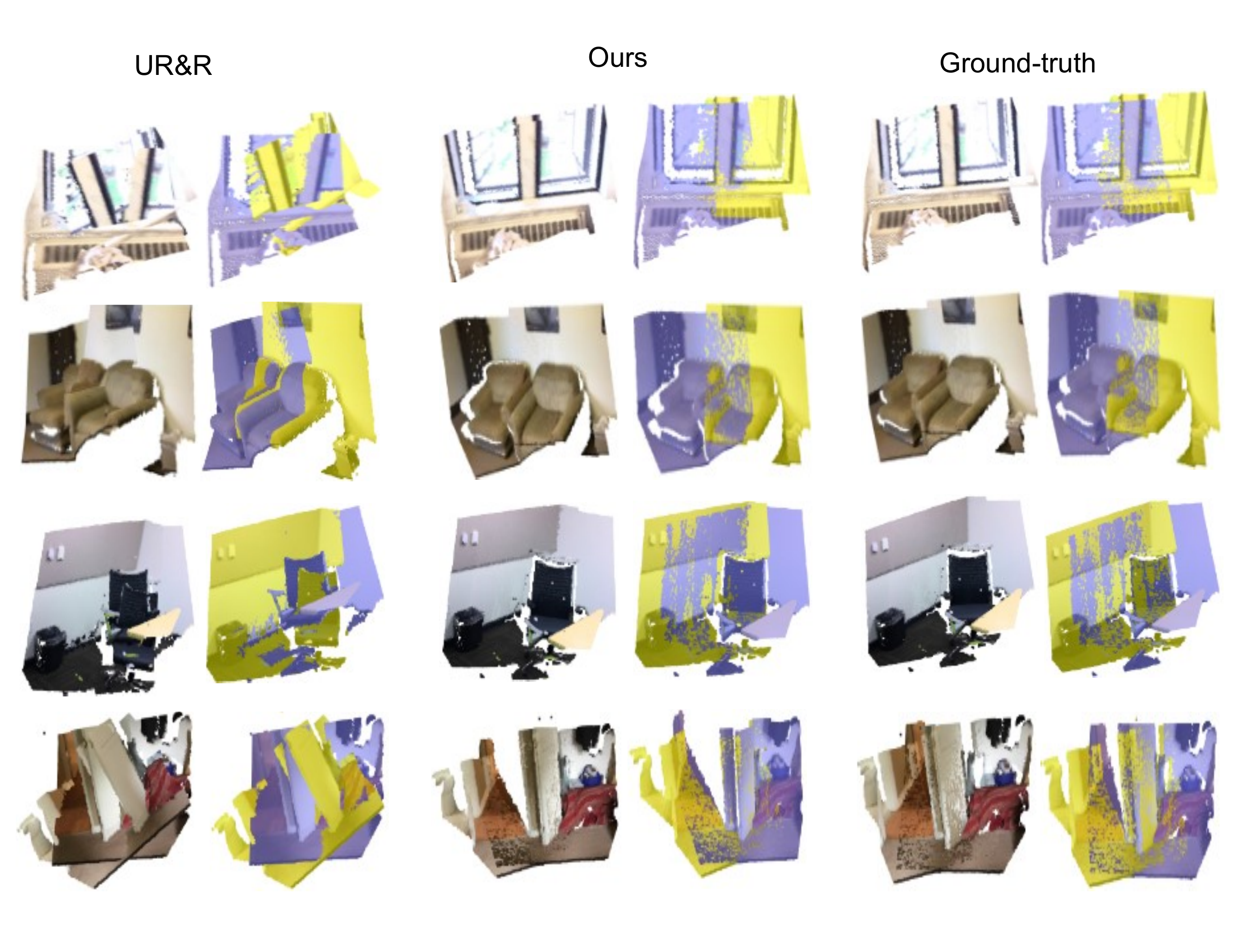}
   \centering
   \caption{Visualization of the point cloud registration results on the ScanNet \cite{dai2017scannet} test dataset by using UR\&R (RGB-D)~\cite{el2021unsupervisedr} and our method. Both methods are trained on the 3D Match dataset. 
   In the \nth{1}, the \nth{3} and the \nth{5} columns, we show the stitched 3D scenes; while we use purple and yellow points to respectively represent the point clouds from the target and reference viewpoints in the \nth{2}, the \nth{4} and the \nth{6} columns. Best viewed on screen.} 
   \label{fig:visualization}
\end{figure}

\section{Visualization of the Registration Results on ScanNet}
In the main paper, we show some visualization results based on the 3D Match dataset to demonstrate our registration performance.
Here, in Figure~\ref{fig:visualization}, we further provide more visualization results to demonstrate the registration performance from both UR\&R~\cite{el2021unsupervisedr} (RGB-D) and our proposed GAVE method on the ScanNet~\cite{dai2017scannet} dataset.
It is observed that our method can still achieves better registration results than UR\&R~\cite{el2021unsupervisedr} on the 3D Match dataset.
For example, for all rows, our method generates much closer stitching results to the ground-truth, while the results from the UR\&R (RGB-D) method exhibits very different reconstruction results to the ground-truth, which indicates the baseline method UR\&R is less effective than our method.
Overall, it shows that we can also achieve better point cloud registration results than UR\&R (RGB-D) on the ScanNet dataset.

%
%

\end{document}